\documentclass{llncs}

\usepackage{listings}
\usepackage{graphicx}
\usepackage{booktabs}
\usepackage{float}
\PassOptionsToPackage{hyphens}{url}
\usepackage{hyperref}
\usepackage{amsmath}
\usepackage[T1]{fontenc}

\begin{document}

\title{Pun Intended: Multi-Agent Translation of Wordplay with Contrastive Learning and Phonetic-Semantic Embeddings}

\titlerunning{Multi-Agent Translation of Wordplay}

\author{
Russell Taylor\inst{1}\orcidID{0009-0007-0702-2375}
\and
Benjamin Herbert\inst{1}\orcidID{0009-0009-0179-3835}
\and
Michael Sana\inst{1}\orcidID{0009-0007-6608-7347}
}

\authorrunning{R. Taylor et al.}

\institute{
Georgia Institute of Technology, Atlanta, GA 30332, USA\\
\email{\{rdtaylorjr,bherbert6,msana3\}@gatech.edu}
}

\maketitle

\begin{abstract}
Translating wordplay across languages has long challenged both professional translators and machine translation systems. We investigate three approaches to translating puns from English to French by combining large language models with linguistic constraints for wordplay generation.

Our baseline uses a large language model with feedback from a discriminator prompted with positive and negative French examples. Our guided reasoning pipeline uses combined phonetic-semantic embeddings to retrieve lexical candidates for wordplay generation. Finally, our multi-agent framework iteratively evaluates and regenerates candidate translations using specialized feedback.

Moving beyond literal translation, our objective is to preserve the linguistic creativity, ambiguity, and humor of the source-text wordplay rather than simply reproduce its vocabulary. The multi-agent and guided chain-of-thought systems ranked first and second, respectively, in the CLEF JOKER 2025 Task~2 competition under expert human evaluation, despite only modest improvements in BLEU and BERTScore.

These findings suggest that both explicit phonetic-semantic guidance and iterative multi-agent evaluation can improve LLM-based wordplay translation relative to direct discriminator-guided generation, particularly when balancing semantic fidelity, phonetic similarity, and natural target-language expression.

\end{abstract}

\section{Introduction}

Translating puns across languages remains one of the most challenging problems in machine translation because it requires preserving both meaning and wordplay.

Puns rely on semantic ambiguity and linguistic incongruity, whereas language models are primarily trained to model regular linguistic and semantic patterns \cite{baziotis2023automatic}. Such incongruity is central to what makes puns surprising and humorous \cite{aarons2017puns}.

Even when a source-language pun is correctly understood, an equivalent pun rarely exists in the target language. Because homonyms and other forms of wordplay are language-specific, preserving both meaning and humor often requires creating entirely new wordplay rather than translating the original literally.

Generating effective target-language wordplay requires creativity in addition to linguistic and cultural knowledge. Producing original humor is difficult even for humans, and professional translators have long regarded humor translation as one of the most challenging translation tasks \cite{miller2019punsters,low2011translating}.

Large language models possess broad linguistic and cultural knowledge acquired from web-scale corpora, making it possible to revisit computational approaches to pun translation that were previously impractical. At the same time, recent advances in computational pun generation provide new mechanisms for modeling semantic ambiguity, phonetic similarity, and iterative refinement.

We make two primary contributions. First, we show that prioritizing functional equivalence over literal correspondence produces stronger human evaluation for cross-lingual wordplay translation. Second, we investigate three progressively more sophisticated LLM-based approaches for achieving this objective: discriminator-guided generation, explicit phonetic-semantic guidance, and iterative multi-agent evaluation.

\section{Related Work}

Computational wordplay translation lies at the intersection of linguistic humor theory, translation studies, machine translation, computational pun generation, and evaluation of creative language.

\subsection{Linguistics of Humor}

The General Theory of Verbal Humor (GTVH) identifies script opposition as a central component of verbal humor, in which incompatible frames of reference are juxtaposed to produce surprise \cite{attardo1991script}. Veisbergs argues that humor can often be preserved in translation even when the underlying linguistic mechanisms change, provided the script opposition remains intact \cite{veisbergs1997contextual}.

Relevance Theory argues that successful pun translation depends on reproducing comparable surprise and cognitive reinterpretation rather than identical lexical forms \cite{yus2003humor}. Aarons further argues that puns rely on tacit linguistic knowledge shared between speaker and listener, suggesting that models with broad implicit linguistic knowledge may be particularly well suited to this task \cite{aarons2017puns}.

\subsection{Professional Human Translation of Wordplay}

Professional translation studies have long recognized wordplay as one of the most difficult forms of translation. Delabastita's influential framework identified eight strategies for translating puns across languages \cite{delabastita1996introduction}, while Low proposed an iterative search procedure that progressively expands semantic and phonetic alternatives until suitable target-language wordplay is found \cite{low2011translating}. Although developed for human translators, Low's framework closely resembles modern retrieval-and-generation pipelines.

\subsection{Machine Translation of Wordplay}

Results from successive CLEF JOKER shared tasks have shown that pun translation remains significantly more difficult than conventional machine translation \cite{ermakova2023overview}. Miller proposed a modular pipeline consisting of pun detection, semantic interpretation, phonetic retrieval, generation, and ranking \cite{miller2019punsters}. Although difficult to implement before modern LLMs, recent reasoning models make this modular architecture practical.

\subsection{Wordplay Generation}

Prompt-based approaches improve wordplay generation through structured reasoning. Xu et al.\ demonstrate that prompt design substantially affects pun recognition, explanation, and generation, while Zhong et al.\ and Wang et al.\ extend these ideas through leap-of-thought prompting and multi-agent generation \cite{xu2024a,zhong2024lets,wang2024innovative}.

A second line of work incorporates explicit linguistic structure. He et al.\ model contextual support for multiple meanings, while Zeng et al.\ improve generation through contrastive learning and discriminator-guided semantic planning \cite{he2019pun,zeng2024barking}. Sharma et al.\ develop IPA-based phonetic embeddings and demonstrate their effectiveness on the English JOKER dataset \cite{sharma2021phonetic}. We build on this methodology by constructing French phonetic embeddings and incorporating them into a guided retrieval framework for cross-lingual pun translation.

\subsection{Evaluation of Generated Wordplay}

Evaluating translated wordplay remains challenging because conventional machine translation metrics such as BLEU and BERTScore reward lexical similarity rather than preservation of figurative meaning and humor. Recent work has therefore proposed evaluation methods tailored to creative language, including metaphor-aware evaluation \cite{wang2024mmte}, literal translation error detection \cite{baziotis2023automatic}, and LLM-based ensemble evaluation of humor using multiple evaluator personas \cite{goes2022crowd}. Our evaluation framework draws on these complementary approaches.

Collectively, these studies suggest that successful wordplay translation requires both linguistic reasoning and evaluation methods that extend beyond lexical similarity.

\section{Methodology}

This section describes the datasets, models, and experimental procedures used in our study. We first present our baseline system and then introduce two progressively more sophisticated approaches that incorporate guided reasoning and iterative multi-agent refinement.

\subsection{Data and Resources}

Our primary dataset was the CLEF JOKER 2025 Task 2 English--French Wordplay Translation dataset \cite{ermakova2025overview,ermakova2025overview-task2}. The training set contains 1,405 English puns paired with 5,838 French translations, while the hidden test set consists of 376 English puns evaluated through the shared task.

We also used the CLEF JOKER 2023 Pun Location and Interpretation dataset \cite{ermakova2023overview}, supplementing it with manual annotations of the pun word, pun type, intended meanings, and supporting context for each English example. The annotations were produced collaboratively by the authors. We then compared them with LLM predictions, manually reviewing any disagreements before finalizing the annotations. These annotations served as the reference standard for pun identification.

For contrastive discrimination, we constructed a balanced dataset by generating one non-pun for each reference French pun in the JOKER dataset. We used \nolinkurl{gemini-2.5-flash-preview-05-20}, regenerating candidates until an independent prompt verified that no wordplay remained.

For phonetic-semantic retrieval, we trained French phonetic embeddings following Sharma et al.\ \cite{sharma2021phonetic} using the Lexique pronunciation database and PanPhon articulatory features \cite{panphon2016}, then concatenated them with pre-trained French FastText semantic embeddings \cite{grave2018learning}.

Gemini 2.5 Pro was used for pun identification, o4-mini for pun-word and synonym translation, \texttt{Lajavaness/bilingual-embedding-large} for bilingual semantic similarity, Mistral Medium 2505 and o4-mini for baseline generation, and Gemini 2.5 Flash as the discriminators. We did not override the providers' default generation parameters.

\subsection{Baseline Pun Generation with Contrastive Discrimination}

\begin{figure}[H]
    \centering
    \includegraphics[scale=0.1]{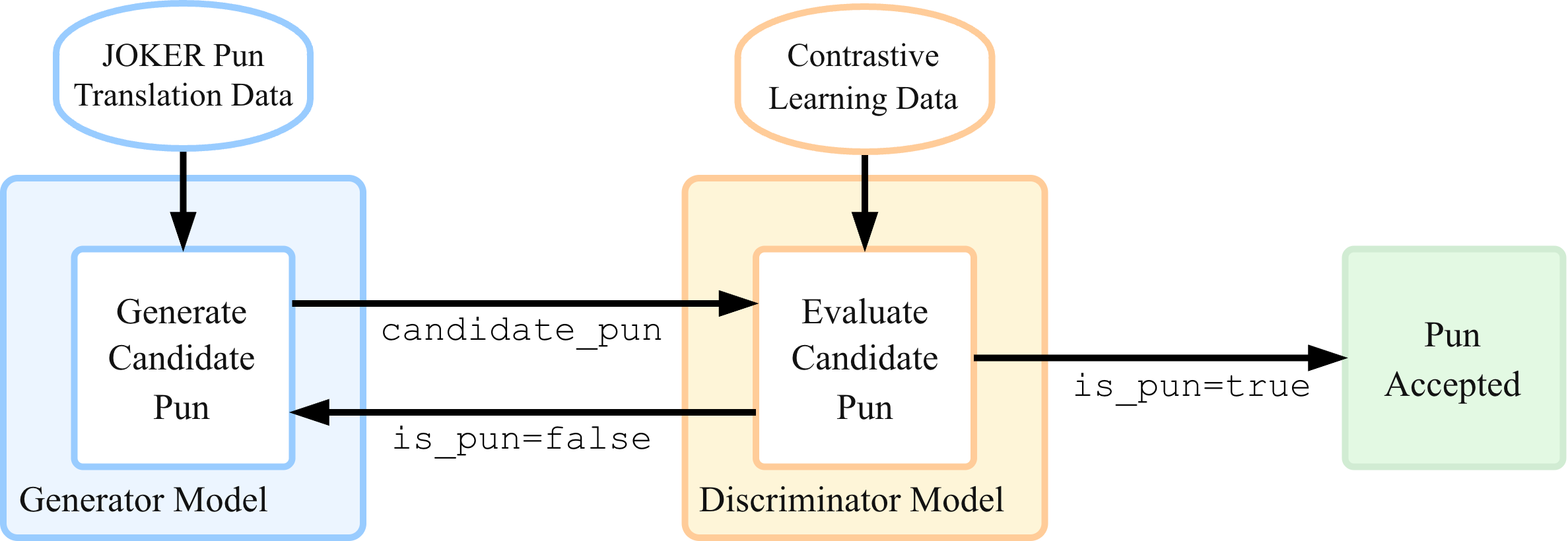}
    \caption{Baseline pun generation pipeline with contrastive discrimination.}
    \label{fig:contrastive}
\end{figure}

Figure~\ref{fig:contrastive} summarizes our baseline pun generation pipeline. Inputs were normalized for punctuation, capitalization, hashtags, and named entities.

Rather than asking models to \emph{translate} the source sentence, we prompted large language models to generate a French pun that preserved both its semantic content and humorous effect. Following Mittal et al.\ \cite{mittal2022ambipun}, prompts encouraged the use of homonyms whose two senses were both supported by the surrounding context.

Generated translations were evaluated by a \texttt{gemini-2.5-flash-preview-05-20} discriminator. We chose a different model from the generator models to reduce potential model-specific bias. Following Zeng et al.~\cite{zeng2024barking}, we used few-shot prompting with 25 positive and 25 negative examples sampled from our balanced dataset of French puns and non-puns. Candidates classified as non-puns were regenerated for up to ten iterations.

\subsection{Guided Chain-of-Thought with Phonetic-Semantic Embeddings}

\begin{figure}[H]
    \centering
    \includegraphics[scale=0.1]{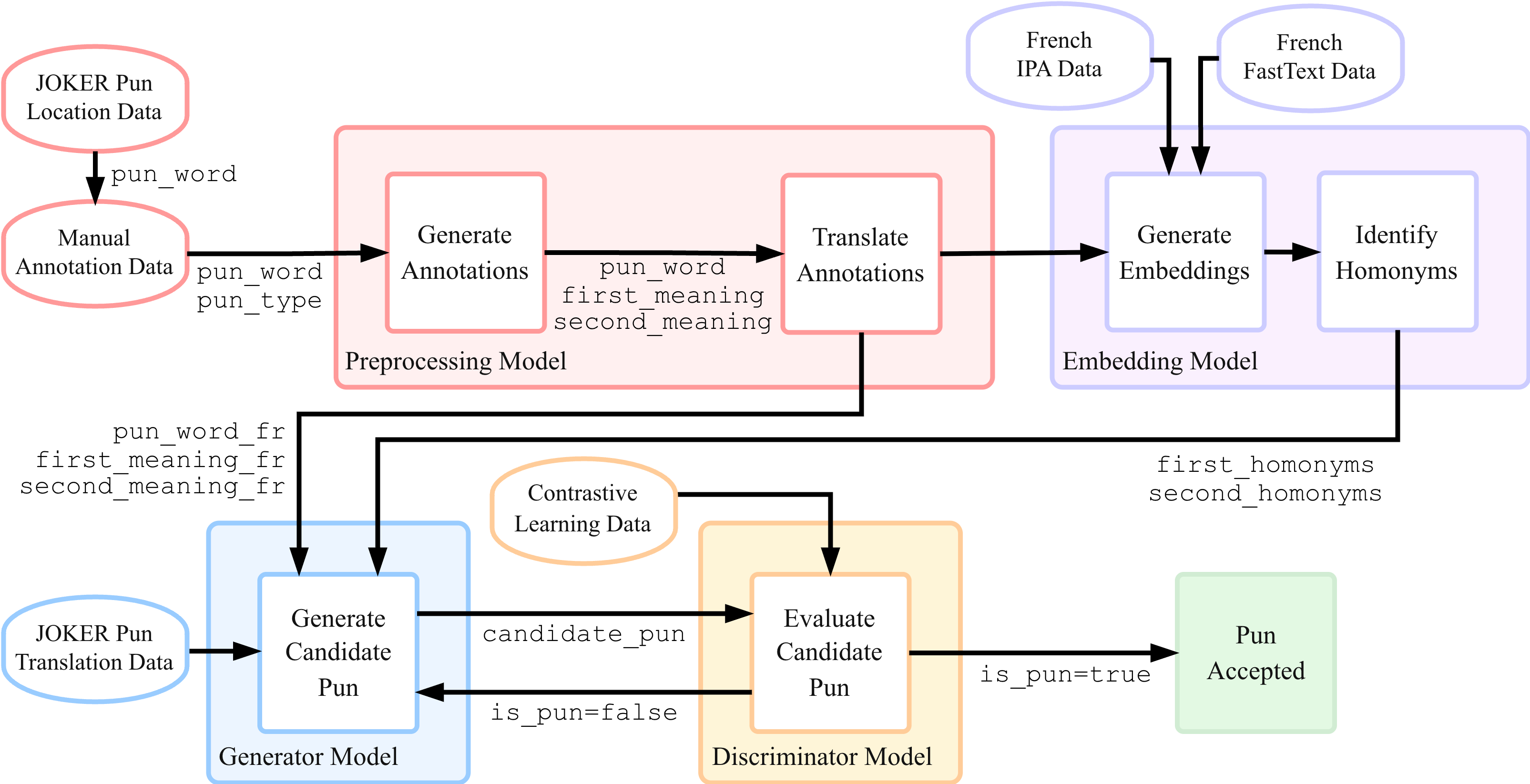}
    \caption{Guided chain-of-thought pipeline with phonetic-semantic embeddings.}
    \label{fig:embedding}
\end{figure}

Figure~\ref{fig:embedding} summarizes our second pipeline. The system identifies the English pun and its two meanings, generates two synonym lists representing each meaning, translates the pun word and synonym lists into French, retrieves French words that are semantically related to one meaning and phonetically related to the other, and supplies the resulting candidates to an LLM for guided generation.

\subsubsection{Pun Word Identification and Translation}

For each English sentence, we used an LLM to identify the pun word, classify the wordplay as homographic or homophonic, and generate synonym lists representing each of its two intended meanings. We then translated the pun word and synonym lists into French.

We evaluated pun identification against our manual annotations and measured translation quality using bilingual embedding cosine similarity. A second prompt assessed whether the translated pun word remained a homonym whose senses aligned with the translated meanings.

\subsubsection{Phonetic-Semantic Retrieval}

We adapted the phonetic embedding method of Sharma et al.~\cite{sharma2021phonetic} to French by representing Lexique pronunciations as IPA sequences and converting them to PanPhon articulatory-feature bigrams \cite{panphon2016}. Similarity between phoneme bigrams was computed using Jaccard similarity:

\[
S((P_{a1},P_{a2}),(P_{b1},P_{b2})) =
\frac{
    \left|F(P_{a1},P_{a2}) \cap F(P_{b1},P_{b2})\right|
}{
    \left|F(P_{a1},P_{a2}) \cup F(P_{b1},P_{b2})\right|
}.
\tag{1}
\]

We used these similarities to train 300-dimensional BiLSTM phonetic embeddings, which were concatenated with 300-dimensional FastText semantic embeddings for joint retrieval. This combination allows retrieval to preserve one intended meaning semantically while identifying a phonetically compatible expression for the other. Following Low \cite{low2011translating}, we retrieved French words that were semantically similar to one translated meaning and phonetically similar to the other. We performed this search in both directions, alternately using each intended meaning as the semantic query and the other as the phonetic query.

We determined empirically that retaining the top two candidates with cosine similarities above 0.75 balanced retrieval quality and noise:

\[
\cos(\vec{w}_{\mathrm{sem}},\vec{S}) > 0.75
\quad\text{and}\quad
\cos(\vec{w}_{\mathrm{phon}},\vec{P}) > 0.75,
\tag{2}
\]

where \(\vec{w}_{\mathrm{sem}}\) and \(\vec{w}_{\mathrm{phon}}\) denote the semantic and phonetic components of a candidate word, and \(\vec{S}\) and \(\vec{P}\) are the corresponding query vectors.

\subsubsection{Guided Pun Generation}

Generation followed one of three strategies according to the identified pun type and the suitability of the translated pun word:

\begin{enumerate}
    \item For homographic puns whose translated pun word remained a French homonym covering both intended meanings, the model was instructed to construct the translation around that word.
    \item For other homographic puns, the model was instructed to select a French homonym whose two senses best approximated the translated meanings.
    \item For homophonic puns, the model was instructed to generate a pair of phonetically similar French words corresponding to the two translated meanings.
\end{enumerate}

Each prompt included the identified pun type, the translated synonym lists, and the retrieved phonetic-semantic candidates.

\subsection{Multi-Agent Evaluation and Refinement}

\begin{figure}[H]
    \centering
    \includegraphics[scale=0.1]{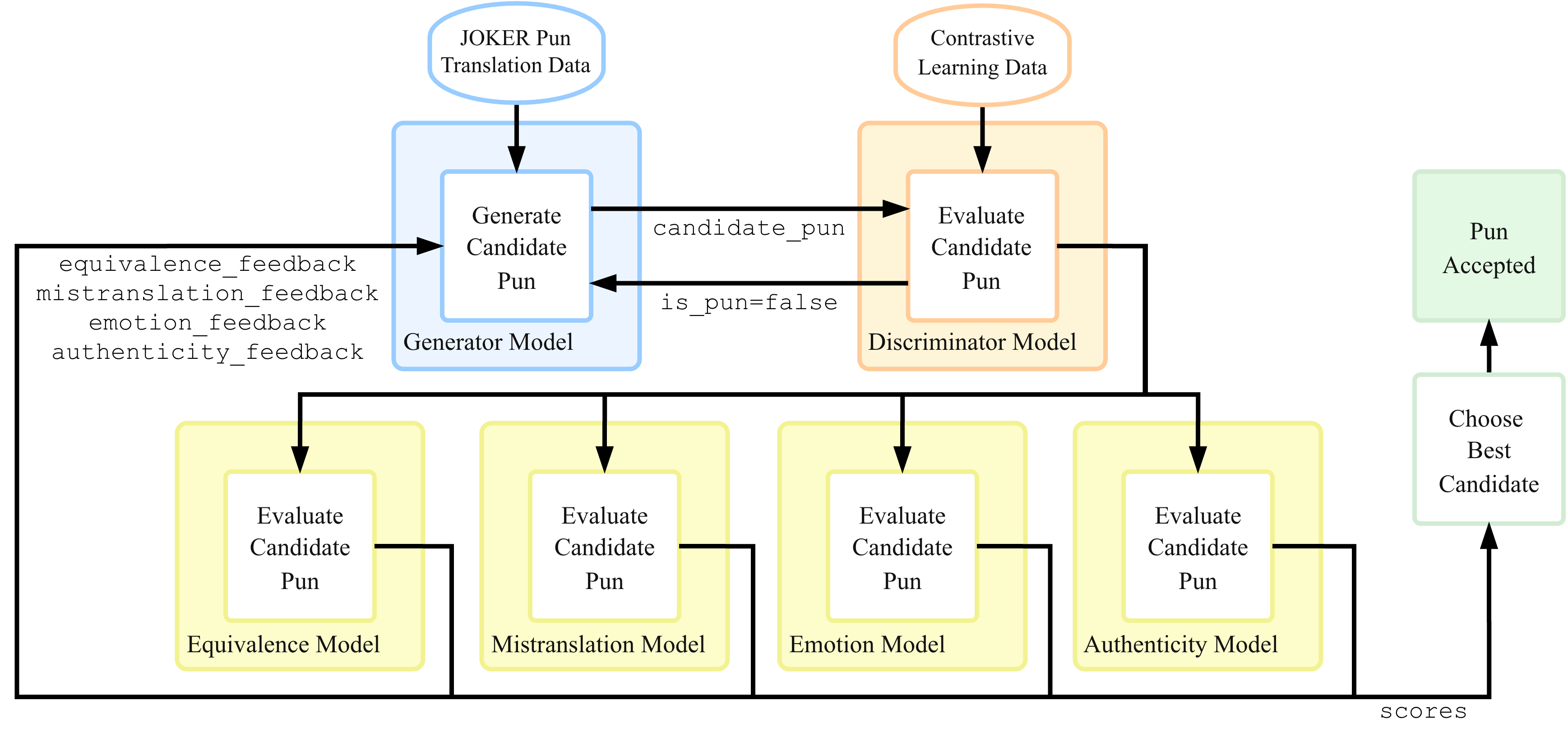}
    \caption{Iterative multi-agent evaluation and refinement pipeline.}
    \label{fig:multiagent}
\end{figure}

Figure~\ref{fig:multiagent} illustrates the iterative refinement framework. Rather than selecting the initial translation directly, specialized LLM evaluators iteratively assessed complementary aspects of translation quality and provided feedback for revision. Following prior work on role-based prompting \cite{shanahan2023roleplaylargelanguagemodels}, we designed four evaluation prompts based on the MMTE dimensions of equivalence, quality, emotion, and authenticity \cite{wang2024mmte}.

Each evaluator assigned a task-specific numerical score together with concise textual feedback: equivalence and quality were rated on 0--2 scales, emotion on a binary 0--1 scale indicating whether emotional content was preserved, and authenticity on a 0--4 naturalness scale. The four scores were averaged, and the resulting feedback was used to generate a revised translation. Refinement continued until the average score reached 2.0 or five iterations had been completed, after which the highest-scoring candidate was retained.

Our code, full prompts, and augmented data are publicly available at \url{https://github.com/dsgt-arc/joker-2025}.
\section{Results}

We first evaluate the intermediate components of the translation pipeline before comparing the three end-to-end translation approaches.

\subsection{Evaluation of Pipeline Components}

Our translation pipeline relies on three intermediate capabilities: identifying the source-language pun and its type, translating the intended meanings into French, and distinguishing successful French wordplay from literal or unsuccessful translations. We therefore evaluated each component independently on the training set before integrating them into the complete system.

\subsubsection{Pun Identification}

Table~\ref{tab:identify} compares automatic identification of the pun location and pun type against our manually annotated reference dataset. Large language models achieved high performance on both tasks, with pun location consistently easier than pun type classification. Pun type classification remained more challenging, likely reflecting ambiguity between homographic and homophonic wordplay.

\begin{table}[H]
\centering
\caption{Performance of LLMs on pun identification compared to manual annotations.}
\label{tab:identify}
\begin{tabular}{lcccccccc}
\toprule
& \multicolumn{4}{c}{\textbf{Pun Location}} &
\multicolumn{4}{c}{\textbf{Pun Type}} \\
\cmidrule(lr){2-5}
\cmidrule(lr){6-9}
\textbf{Model}
& \textbf{Acc.}
& \textbf{Prec.}
& \textbf{Rec.}
& \textbf{F1}
& \textbf{Acc.}
& \textbf{Prec.}
& \textbf{Rec.}
& \textbf{F1} \\
\midrule
Gemini 2.5 Pro   & 0.97 & 1.00 & 0.97 & 0.97 & 0.89 & 0.90 & 0.89 & 0.89 \\
o3               & 0.95 & 0.99 & 0.95 & 0.95 & 0.88 & 0.90 & 0.88 & 0.88 \\
o4-mini          & 0.93 & 0.99 & 0.93 & 0.93 & 0.88 & 0.89 & 0.88 & 0.88 \\
Claude Sonnet 4  & 0.93 & 0.99 & 0.93 & 0.92 & 0.86 & 0.88 & 0.86 & 0.86 \\
Gemini 2.5 Flash & 0.89 & 0.99 & 0.89 & 0.89 & 0.87 & 0.90 & 0.87 & 0.87 \\
GPT-4.1          & 0.88 & 0.99 & 0.88 & 0.88 & 0.82 & 0.86 & 0.82 & 0.82 \\
\bottomrule
\end{tabular}
\end{table}

\subsubsection{Translation of Synonym Lists}

Table~\ref{tab:translate} compares several large language models for translating the synonym lists representing the two intended meanings of each pun. The strongest models performed similarly, so we selected o4-mini because of its lower cost and latency. All evaluated LLMs also outperformed Google Translate on this constrained translation task.

\begin{table}[H]
\centering
\caption{Quality of synonym list translation measured using bilingual English--French embeddings.}
\label{tab:translate}
\begin{tabular}{lccc}
\toprule
\textbf{Model}
& \textbf{Mean Cosine}
& \textbf{Variance}
& \textbf{Error Rate} \\
\midrule
Gemini 2.5 Pro      & 0.82 & 0.01 & 0.01 \\
o4-mini             & 0.81 & 0.01 & 0.00 \\
Claude Sonnet 4     & 0.81 & 0.01 & 0.01 \\
Mistral Medium 2505 & 0.79 & 0.01 & 0.02 \\
Gemini 2.5 Flash    & 0.79 & 0.02 & 0.18 \\
Google Translate    & 0.78 & 0.01 & 0.00 \\
\bottomrule
\end{tabular}
\end{table}

\subsubsection{Contrastive Discriminator}

When tested using a manually labeled evaluation set of 450 examples, our contrastive discriminator achieved 100.0\% accuracy on negative examples and 99.1\% accuracy on positive examples. These results indicate that the discriminator can provide a reliable feedback signal for iterative generation.

\subsection{End-to-End Translation Performance}

Having validated the individual pipeline components, we next evaluated the complete translation systems on the official CLEF JOKER 2025 Task~2 test set. Performance was measured using BLEU, BERTScore, the shared-task pun location metric, and manual evaluation. The shared-task pun location metric measures whether the generated translation places its wordplay at the same location as one of the reference translations. Manual evaluation was performed by a native French speaker with graduate-level training in wordplay translation. A translation was considered successful if it preserved the source meaning (fully or partially) while producing target-language wordplay \cite{ermakova2025overview-task2}.

Table~\ref{tab:end_to_end} summarizes the official evaluation results. The multi-agent system ranked first among the 51 submitted systems under both the pun location metric and manual evaluation, while the guided system ranked second on both measures. The multi-agent system correctly localized the reference pun in 156 of the 1,682 evaluated translations, compared with 132 for the guided system and 60 for the baseline. Under manual evaluation, the three systems produced successful translations for 37, 36, and 20 of the 42 evaluated examples \cite{ermakova2025overview-task2}.

\begin{table}[t]
\centering
\caption{Official evaluation results on the CLEF JOKER 2025 Task~2 test set. Rank among 51 official submissions is shown in parentheses.}
\label{tab:end_to_end}
\small
\begin{tabular}{lrrrr}
\toprule
\textbf{System}
& \textbf{BLEU}
& \textbf{BERTScore}
& \textbf{Pun Location}
& \textbf{Manual} \\
\midrule
Multi-Agent
& 21.41 (41)
& 80.66 (41)
& \textbf{9.27\% (1)}
& \textbf{88.09\% (1)} \\

Guided CoT
& 16.52 (45)
& 78.42 (45)
& 7.85\% (2)
& 85.71\% (2) \\

Baseline
& 14.94 (46)
& 78.30 (46)
& 3.57\% (45)
& 47.62\% (25) \\
\bottomrule
\end{tabular}
\end{table}

\section{Discussion} 

Our experiments yielded several lessons about pun translation, evaluation, and system design. Some confirmed long-standing ideas from translation theory, while others challenged common standards in computational translation.

\subsection{What We Learned About Pun Translation} 

Our results reinforce a simple idea: pun translation is about recreating humor, not reproducing text.

\subsubsection{Functional Equivalence Beats Lexical Correspondence}

Translation theory has long argued that successful pun translation depends on preserving the joke rather than the words. This principle guided the design of our systems from the outset. As Low writes, ``If a joke is not translated as a joke, the translation is bad'' \cite{low2011translating}. We therefore designed our systems to pursue functional equivalence rather than lexical correspondence, encouraging them to recreate the joke rather than translate it literally.

For example, rather than literally translating pun en\_160, ``That makes 144,'' said Tom, grossly, our system produced «~Ça fait douze au carré~», dit-il carrément, replacing the English wordplay with a different French pun that preserves the joke.

Throughout development, we observed that LLMs default to lexical and semantic equivalence when asked to translate. While desirable for conventional machine translation, this behavior is often at odds with pun translation. We therefore encouraged the models to recreate the joke rather than translate it literally.

Our results provide empirical support for this long-standing principle of translation theory. Although our systems departed from conventional translation objectives, they achieved the strongest human evaluation results in the JOKER shared task. We believe our emphasis on functional equivalence was a major contributor to this success.

\subsubsection{Automatic Metrics Fundamentally Misunderstand Pun Translation}

Our results suggest that current automatic metrics reward the wrong objective for pun translation. BLEU and BERTScore reward lexical correspondence, whereas successful pun translation often requires abandoning it in order to preserve the joke. We deliberately optimized for functional equivalence rather than lexical overlap.

The result was a striking inversion of the leaderboard: our systems ranked near the bottom according to BLEU and BERTScore, yet achieved the strongest human evaluation results in the shared task. This suggests that lexical-overlap metrics systematically undervalue successful pun translation.

We therefore suggest that future evaluation criteria should  prioritize functional equivalence over lexical correspondence in computational pun translation.

\subsubsection{Puns Resist Simple Taxonomies}

Building computational systems made one fact unmistakable: puns are far more diverse than simple taxonomies suggest.

Our initial distinction between homophonic and homographic puns quickly proved inadequate. Many puns combine multiple forms of ambiguity or contain multiple pun words. For example, pun en\_3987, ``My mate who's an origami teacher has quit her job...'', simultaneously relies on \emph{paperwork}, \emph{folding under pressure}, and \emph{couldn't cut it}. These observations suggest that future computational approaches would benefit from richer, multidimensional taxonomies of wordplay grounded in long-established linguistic frameworks \cite{HempelmannMiller2022,Delabastita1996,AttardoRaskin1991,Sun2022ExPUN}.

\subsubsection{Delabastita Was Right}

Our findings provide compelling support for Delabastita's optimism about pun translation. Delabastita argues that ``excellent translation solutions can be found for many puns, if only translators use to the full the linguistic resources and textual leeway available to them in recreating the pragmatic function of the original wordplay'' \cite{delabastita1994}.

We began this paper by emphasizing the extraordinary difficulty of pun translation \cite{ermakova2025overview}. Our results do not diminish that challenge, but they do show that it is far from insurmountable. Expert human evaluation judged 88.09\% of our sampled translations to be successful, providing strong empirical evidence that high-quality computational pun translation is not merely possible, but achievable.

\subsection{What We Learned About Our Methodology} 

Our experiments also yielded several methodological lessons about prompting, supervision, evaluation, and system architecture.

\subsubsection{Creativity Before Constraint}

Our systems were almost entirely unsupervised, despite the shared task providing a purpose-built training set and many competing systems relying heavily upon it. Rather than constraining generation from the outset, they generated candidate translations before refining them through guidance and iterative evaluation. This is consistent with the central intuition behind \emph{Leap-of-Thought} \cite{zhong2024lets}: preserving creative freedom before introducing additional constraints.

These observations suggest two broader lessons. First, creativity appears to benefit more from guidance than prescription. Rather than attempting to dictate how a pun should be translated, our most successful systems encouraged exploration and then evaluated the results. Second, computational pun translation is fundamentally a search problem. This gradually changed how we thought about the task. Rather than teaching a model how to translate puns, we found ourselves helping it search for translations that already lay within its creative repertoire.

\subsubsection{Architecture Shapes Translation Strategy}

System architecture shaped how our models translated puns, not just how well they translated them.

While evaluating our final outputs, we were not surprised to discover that our systems often produced different translations. What surprised us was that they consistently adopted different translation strategies. Our multi-agent system tended to preserve both the linguistic structure and humorous effect of the source through iterative refinement, while the guided chain-of-thought system remained more closely anchored to the English through explicit linguistic reasoning. By contrast, our baseline frequently ignored the English altogether, producing fluent French puns, sometimes even well-known ones, with little lexical or semantic correspondence to the source.

For pun en\_40, ``The class took a field trip to a meat processing plant, but what they saw was just offal,'' the multi-agent system produced des abats-ominables, recreating both the setting and the wordplay, whereas the guided system produced the more literal Un vrai trip de tripes!. The baseline instead abandoned the original joke entirely, replacing it with an unrelated French pun. These patterns occurred quite often.

Our experiments suggest that architecture is not merely an implementation detail: it influences the kinds of translation strategies that LLMs are likely to adopt. In our systems, prompting and agent interactions encouraged progressively stronger adherence to functional equivalence, whereas the unconstrained baseline often treated the source text as inspiration rather than something to translate.

\subsubsection{LLMs Have Become Surprisingly Good Evaluators}

One of our biggest surprises was how effective simple LLM-based evaluation proved to be.

Our two best systems achieved virtually identical human evaluation scores within one point of each other, a statistically insignificant difference on such a small sample. Nevertheless, it was the multi-agent evaluator that came out ahead. This surprised us because our guided chain-of-thought system required substantially more engineering, and we had expected that the additional linguistic reasoning would prove decisive. Instead, relatively simple run-time LLM evaluation was sufficient to match and slightly outperform it.

These results may also provide a promising alternative to traditional evaluation metrics. If lexical-overlap metrics optimize the wrong objective for pun translation, LLM-based evaluators may offer a more suitable way to assess whether a translation succeeds as a joke. As these evaluators continue to improve, they may become an increasingly valuable tool for computational humor research.

\subsubsection{Low's Polygonal Algorithm Is the Way Forward} 

Our retrieval experiments convinced us that Low's polygonal algorithm remains a promising roadmap for computational pun translation. We set out to implement Low's proposal, but ultimately implemented only its first stage. Rather than exhausting the idea, we came away convinced that we had barely begun.

In retrospect, Low's framework reads less like translation theory than like an outline of a computational algorithm, and we believe it provides a compelling foundation for future work in computational pun translation \cite{low2011translating}.

\section{Conclusions}

This study presented a multi-stage methodology for translating puns by prioritizing the preservation of humor and wordplay over literal equivalence. Our approach progressed from baseline prompting to guided chain-of-thought reasoning using phonetic-semantic embeddings and finally to a multi-agent refinement loop.

Our results showed that large language models are capable of nuanced linguistic reasoning, and that improvements to both pun generation and run-time evaluation contribute to translation quality. Ultimately, our findings suggest that successful computational pun translation depends less on reproducing the original words than on recreating their humorous effect.

\begin{credits}

\subsubsection{\ackname}

This work received financial support from DS@GT ARC. We thank Anthony Miyaguchi and Murilo Gustinelli for organizing and leading the DS@GT ARC program, and for their guidance, support, and insightful discussions throughout this project. We also thank the CLEF JOKER 2025 organizers for developing and maintaining the shared task and evaluation framework, and the anonymous reviewers for their constructive feedback, which helped improve this paper.

\subsubsection{\discintname}

The authors declare that they have no competing interests.

\end{credits}

\bibliographystyle{splncs04}
\bibliography{main}

\end{document}